\documentclass[conference]{IEEEtran}
\IEEEoverridecommandlockouts
\usepackage{cite}
\usepackage{xcolor}
\usepackage{amsmath,amssymb,amsfonts}
\usepackage{graphicx}
\usepackage{textcomp}
\usepackage{amsmath}
\usepackage[ruled,vlined]{algorithm2e}
\usepackage[noend]{algpseudocode}
\usepackage{float}
\usepackage{hyperref}
\usepackage{amsthm}

\newcommand{\pluseq}{\mathrel{+}=}

\def\BibTeX{{\rm B\kern-.05em{\sc i\kern-.025em b}\kern-.08em
    T\kern-.1667em\lower.7ex\hbox{E}\kern-.125emX}}
\begin{document}
\title{Ordinary Differential Equation and Complex Matrix Exponential  for Multi-resolution Image Registration}
\author{Abhishek Nan, Matthew Tennant, Uriel Rubin and Nilanjan Ray
\thanks{This work was supported in part by NSERC Discovery Grants.}
\thanks{Abhisked Nan and Nilanjan Ray are with the Department of Computing Science, Univeristy of Alberta, Edmonton, AB T6G2E8, Canada (e-mails: \{anan1, nray1\}@ualberta.ca). }
\thanks{Matthew Tennant is 
with the Department of Ophthalmology, University of Alberta, Edmonton, Alberta, Canada (e-mail: mtennant@ualberta.ca).}
\thanks{Uriel Rubin is with 
the Department of ophthalmology, Hospital Aleman, Buenos Aires, Argentina (e-mail: urielrubin@gmail.com).}}

\maketitle

\begin{abstract}
Autograd-based software packages have recently renewed interest in image registration using homography and other geometric models by gradient descent and optimization, e.g., AirLab \cite{DBLP:journals/corr/abs-1806-09907} and DRMIME \cite{Nan2020}. In this work, we emphasize on using complex matrix exponential (CME) over real matrix exponential to compute transformation matrices. CME is theoretically more suitable and practically provides faster convergence as our experiments show. Further, we demonstrate that the use of ordinary differential equation (ODE) as an optimizable dynamical system can adapt the transformation matrix more accurately to the multi-resolution Gaussian pyramid for image registration. Our experiments include four publicly available benchmark datasets, two of them 2D and the other two being 3D. Experiments demonstrate that our proposed method yields significantly better registration compared to a number of off-the-shelf, popular, state-of-the-art image registration toolboxes. Our software is provided in \href{https://github.com/abnan/ODECME}{GitHub}.
\end{abstract}


\section{Introduction}
\label{sec:introduction}
\IEEEPARstart{I}{mage} registration is the task of finding the correspondences across two or more images. This is often used to tackle problems in the field of medical imaging, remote sensing, etc. For instance, when we want to analyze how the anatomy of a patient's body part changes over time, we need snapshots of it over time. Not only could the source camera change, the location, orientation of the camera as well as anatomy of the patient are variables that could change over time. In such scenarios, doing a comprehensive analysis becomes difficult and hence, registration becomes a prerequisite before any further analysis can be done.

In some cases, there might be different cameras used to capture complimentary information of the same organ at a given time. In such a case, the information from different sources needs to be registered and such a task is called multi-modality image registration. This is done often to provide a more holistic analysis of a subject. For example, MRI (Magnetic Resonance Imaging) scans could be conducted at 1.5T or 3T, with T (Tesla) specifying the strength of the magnet used in the MRI machine. The body's tissues, muscles, fats, etc. all react differently to differing MRI exposures, and this often helps to provide complimentary information by using multiple imaging sources.

There are multiple approaches to image registration, and may be broadly classified into two families based on how the registration parameters are obtained: learning based and optimization based. For instance, Cheng et al. \cite{cheng2018deep} train a binary classifier to learn the correspondence of two image patches and the classification output is transformed to a continuous probability value, which is then used as the similarity score. Alireza et al. \cite{sedghi2018semi} have a similar approach, but rather than needing well aligned training data, they propose a strategy to learn a deep similarity metric from roughly aligned training data. The benefit of such learning based approaches is that once trained on a dataset, inference is quite fast. On the other hand, the drawback of learning based approaches is that they need large amounts of training data to achieve satisfactory results and furthermore, they will not perform well on pairs which are drastically different from the training set. In this work, we utilize an optimization based approach where running time may be longer than the learning based methods, but it is accurate and it does not require any training data.

Often the model used to parametrize the registration parameters are fashioned in a hierarchical manner; i.e. first a global transformation using homography (or it's subsets) is used to register the images as much as possible before applying more elaborate techniques such as deformable registration. Thus, the success of deformable registration methods is highly dependent on how successful the initial registration step was. Our approach is based on optimization to solve for this initial global homographic transform.

Even though several software toolboxes exist for optimization-based image registration using the family of homography transformations, we believe that opportunities still exist for improvement. In this work, we point out that representing transformation matrices using a matrix exponential, especially, complex matrix exponential (CME) leads to faster convergence. CME enjoys a theoretical guarantee that repeated compositions of matrix exponential are not required during optimization, unlike the real case. Furthermore, using a matrix exponential, both the forward and the reverse transformations can be easily added to the registration objective function for a robust design.

In this work, we also point out that a precise design of transformation matrix is possible for the multi-resolution image registration using a dynamical system modeled by a neural network. This dynamical system leads to an initial value ordinary differential equation (ODE) that can adapt a transformation matrix quite accurately to the multi-resolution image pyramids, which are significant for image registration. Our ODE-based framework leads to a more accurate image registration algorithm.

Using the aforementioned two elements, ODE and CME, we present a novel multi-resolution image registration algorithm ODECME that can accommodate both 2D and 3D image registration, mono-modal and multi-modal \cite{MI_survey} cases, and any differentiable loss or objective function including MINE (mutual information neural estimation) \cite{belghazi2018mine}. Our implementation uses PyTorch \cite{NEURIPS2019_9015}, which has a capability of GPU acceleration and automatic gradient computation. Experiments on four publicly available benchmark datasets demonstrate new state-of-the-art performance using ODECME.

\section{Background}

\subsection{Matrix Exponential for Image Registration}
Optimization-based image registration retrieves a transformation matrix $H$ (e.g., homography, affine, rigid body, similarity, etc.) that warps a moving image $M$ to the template image $T$ by optimizing a cost function $D$:
\begin{equation}
    \min_H D(T,Warp(M,H)).
\label{eqn:opt}
\end{equation}
For a differentiable loss function $D$ and a differentiable $Warp$ program, gradient descent can minimize (\ref{eqn:opt}). Representing the transformation matrix $H$ by matrix exponential \cite{schroter2010lie, Nan2020} offers several advantages, e.g., a rigid body transformation matrix can be implicitly represented without any explicit constraints on the elements of $H,$ making the optimization unconstrained.

Using matrix exponential a transformation matrix $H$ is represented by the exponential map or a number of compositions of such maps from suitable matrix Lie algebras to the corresponding matrix Lie groups, such as \textit{SO(3)}, \textit{SE(2)}, etc. \cite{taylor1994minimization, trouve1998diffeomorphisms}. Wachinger \& Navab \cite{wachinger2012simultaneous} show that spatial transformations represented by matrix exponential help because unconstrained optimization can be performed over 3D rigid transformations. Among more recent works, data representations in orientation scores as a function on the Lie group \textit{SE(2)} has been used for template matching\cite{bekkers2017template} with cross-correlation.

As a concrete example, to represent the 2D affine transformations, \textit{Aff(2)} group, the following six generators are used in Lie algebra \cite{Nan2020}:
\begin{equation*}
\begin{split}
& B_1=
\begin{bmatrix}
0 & 0 & 1\\
0 & 0 & 0\\
0 & 0 & 0\\
\end{bmatrix},
B_2=
\begin{bmatrix}
0 & 0 & 0\\
0 & 0 & 1\\
0 & 0 & 0\\
\end{bmatrix},
B_3=
\begin{bmatrix}
0 & 1 & 0\\
0 & 0 & 0\\
0 & 0 & 0\\
\end{bmatrix}, \\
& B_4=
\begin{bmatrix}
0 & 0 & 0\\
1 & 0 & 0\\
0 & 0 & 0\\
\end{bmatrix},
B_5=
\begin{bmatrix}
1 & 0 & 0\\
0 & -1 & 0\\
0 & 0 & 0\\
\end{bmatrix},
B_6=
\begin{bmatrix}
0 & 0 & 0\\
0 & -1 & 0\\
0 & 0 & 1\\
\end{bmatrix}.
\end{split}
\end{equation*}

Using these six generators, an affine transformation matrix can be expressed as $Mexp(\sum_{i=1}^6v_i B_i)$, where $v=[v_1,v_2,...,v_6]$ is a parameter/coefficient vector. $Mexp$ is the matrix exponentiation operation that can be computed by the power series on matrix $B$ \cite{Hall2015},
\begin{equation}
    Mexp(B) = \sum_{n=0}^{\infty} \frac{B^n}{n!},    
\label{eqn:mat_exp}
\end{equation}
which can be truncated after a few terms (e.g., 10) for an accurate enough representation of a transformation matrix \cite{Nan2020}. A more sophisticated algorithm can also be applied for matrix exponential computation \cite{mat_exp}, as long as it is easily differentiable. Using the matrix exponential representation for a transformation matrix, $H=Mexp(\sum_i v_i B_i)$, the image registration optimization (\ref{eqn:opt}) takes the following form:
\begin{equation}
    \begin{split}
    \min_{v_1,v_2,...} & D(T,Warp(M,Mexp(\sum_i v_i B_i))) + \\
    & D(M,Warp(T,Mexp(-\sum_i v_i B_i))),
    \end{split}
\label{eqn:opt_me}
\end{equation}
where we have also added a cost for registering template $T$ to moving image $M,$ making the optimization more robust. The symmetric objective function denotes a clear advantage of matrix exponential, where the inverse transform can be easily added to the differentiable cost function. Thus, we can apply gradient descent by automatic differentiation (i.e., chain rule) to adjust parameters $v_i$.

For 3D data, we can have the \textit{SE(3)} and \textit{Sim(3)} groups. The \textit{SE(3)} group represents all 3D rigid transformations, i.e. it has six degrees of freedom, which are the three axes of rotation and three directions of translation. The six generators \cite{eade2013lie} are:
\begin{equation*}
\begin{split}
& B_1=
\begin{bmatrix}
0 & 0 & 0 & 1\\
0 & 0 & 0 & 0\\
0 & 0 & 0 & 0\\
0 & 0 & 0 & 0\\
\end{bmatrix},
B_2=
\begin{bmatrix}
0 & 0 & 0 & 0\\
0 & 0 & 0 & 1\\
0 & 0 & 0 & 0\\
0 & 0 & 0 & 0\\
\end{bmatrix}, \\
& B_3=
\begin{bmatrix}
0 & 0 & 0 & 0\\
0 & 0 & 0 & 0\\
0 & 0 & 0 & 1\\
0 & 0 & 0 & 0\\
\end{bmatrix}, 
 B_4=
\begin{bmatrix}
0 & 0 & 0 & 0\\
0 & 0 & -1 & 0\\
0 & 1 & 0 & 0\\
0 & 0 & 0 & 0\\
\end{bmatrix},\\
& B_5=
\begin{bmatrix}
0 & 0 & 1 & 0\\
0 & 0 & 0 & 0\\
-1 & 0 & 0 & 0\\
0 & 0 & 0 & 0\\
\end{bmatrix},
B_6=
\begin{bmatrix}
0 & -1 & 0 & 0\\
1 & 0 & 0 & 0\\
0 & 0 & 0 & 0\\
0 & 0 & 0 & 0\\
\end{bmatrix}.
\end{split}
\end{equation*}
The similarity group \textit{Sim(3)} adds another degree of scaling to 3D rigid transformations \textit{SE(3)}. Their generators are the same except for an additional one:
\begin{equation*}
B_7=
\begin{bmatrix}
0 & 0 & 0 & 0\\
0 & 0 & 0 & 0\\
0 & 0 & 0 & 0\\
0 & 0 & 0 & -1\\
\end{bmatrix}.
\end{equation*}

\subsection{Multi-resolution Computation}

Literature on scale-space \cite{Witkin, Lindeberg} has shown that objects and edges have intrinsic scales in an image. Gaussian pyramid has become a standard and discrete method for capturing the continuous scale-space for an image. Image pyramid-based computations are routinely used for motion estimation \cite{szeliski2004image} that can drastically reduce computations by means of a hierarchical search beginning at the top of the image pyramid and ending at the bottom of the pyramid, i.e., the original resolution of the image. Optimization-based optical flow computation also adopts this multi-resolution technique \cite{Ray2011}. Image registration methods have also adopted multi-resolution image pyramids \cite{thevenaz1998pyramid, kruger1998image, alhichri2002multi} that have shown better convergence and accuracy for gradient-based optimizations.

Image resolution and registration cost, such as mutual information have their complex interactions. Irani and Anandan \cite{irani1998robust} have studied that effectiveness of mutual information decreases as one moves towards coarser resolution. Wu and Chung \cite{wu2004multimodal} have combined mutual information and sum of difference (SAD) for multi-modal and multi-resolution registration. 

Sun et al. \cite{sun2013simultaneous} have shown that instead of computing transformations at the coarsest level of the pyramid and propagating it towards the finer levels, one can use all the pyramid levels simultaneously for better convergence and accuracy. In our recent study \cite{Nan2020}, we have also noted that adding cost of registration simultaneously for all pyramid levels and optimizing the combined cost function is more beneficial.

The registration optimization problem (\ref{eqn:opt_me}) using the multi-resolution approach takes the following form:
\begin{equation}
    \begin{split}
    \min_{v_1,v_2,...} \sum_{l=1}^L & \{D(T_l,Warp(M_l,Mexp(\sum_i v_i B_i))) + \\
    & D(M_l,Warp(T_l,Mexp(-\sum_i v_i B_i)))\},
    \end{split}
\label{eqn:opt_me_mr}
\end{equation}
where $T_l$ and $M_l$ for $l=1,...,L,$ are two image pyramids, with $L$ being the coarsest/maximum level in the pyramid. $T_1=T$ and $M_1=M$ are the original template and moving images, respectively. This formulation assumes that image interpolation $Warp$ and transformation matrices, $Mexp(\sum_i v_i B_i)$ and $Mexp(-\sum_i v_i B_i)$, use the same range of pixel coordinates, such as the canonical range, $[-1,1]\times[-1,1]$ for all resolutions.

\subsection{Image Registration Metrics}

For gradient descent (or ascent)-based optimization, image registration requires a differentiable loss/cost/objective function. Mean squared error (MSE) \cite{MSE} and normalized cross correlation (NCC) \cite{NCC} are two widely used cost functions. While MSE is not suitable for multi-modal image registration, NCC can often serve as an objective here. More specialized measures for multi-modal image registration includes mutual information (MI) \cite{JHMI} and normalized mutual information (NMI) \cite{NMI}. However, not all forms of MI are easily differentiable and for multi-channel images, MI computation may not be trivial. Recently, a differentiable form for mutual information called mutual information neural estimation (MINE) \cite{belghazi2018mine} has been proven quite successful for image registration \cite{Nan2020}.

MINE between two images $P$ and $Q$ is defineed as follows \cite{belghazi2018mine}:
\begin{equation}
    \begin{split}
    MINE(P,Q)= & \frac{1}{N}\sum_{i}f_\theta(P_{I_i}, Q_{I_i}) - \\
    & log(\frac{1}{N}\sum_{i}exp(f_\theta(P_{I_i}, Q_{I^{rp}_i}))),
    \end{split}
\label{eqn:MINE}
\end{equation}
where $I$ denotes a randomly sampled set of $N$ pixel locations. The set $I^{rp}$ denotes a random permutation of the set $I.$ $I_i$ and $I^{rp}_i$ denote $i^\text{th}$ element (i.e., pixel location) of the sets $I$ and $I^{rp},$ respectively. $f_\theta$ is a fully connected neural network with parameters $\theta$ \cite{Nan2020}.

\section{Proposed Method}

\subsection{ODE for Multi-resolution Image Registration}
Image structures are slightly shifted through multi-resolution Gaussian image pyramids \cite{Witkin}. So, a transformation matrix suitable for a coarse resolution may need a slight correction when used for a finer resolution. To mitigate this issue, we model matrix exponential parameters as a continuous function $v(s)$ of resolution $s.$ The change in $v(s)$ over resolution $s$ can be modeled by a neural network $g_\phi$ with parameters $\phi$:
\begin{equation}
\frac{dv(s)}{ds} = g_\phi(s,v(s)).
\label{eq:ode}
\end{equation}
Using Euler method \cite{Butcher2016} the ordinary differential equation (ODE) (\ref{eq:ode}) can be solved for all resolution levels $1, 2, ..., L$:
\begin{equation}
    \begin{split}
    v_L & = u\\
    \text{for}~ l&=L-1,L-2,\cdots,1\\
    v_l & = v_{l+1} +  (s_l-s_{l+1})g_\phi(s_{l+1},v_{l+1}),
    \end{split}
\label{eq:euler}
\end{equation}
where $v_l = [v_{l,1},v_{l,2},...]$ are the matrix exponential coefficients for resolution level $l$ and $s_l = d^{-l+1},~l=1,2,..,L$ denote the discrete resolutions in powers of the downscale factor $d$. $u=[u_1,u_2,...]$ is the initial value vector in the ODE and it is an optimizable parameter of the model along with the neural network parameters $\phi.$ We have also used 4-point Runge-Kutta method (RK4) \cite{Butcher2016} for the above recursion:
\begin{equation}
    \begin{split}
    v_L & = u\\
    \text{for}~ l&=L-1,L-2,\cdots,1\\
    h & = s_l-s_{l+1}, \\
    k_1 & = hg_\phi(s_{l+1},v_{l+1}),\\
    k_2 & = h g_\phi(s_{l+1}+\tfrac{1}{3}h,v_{l+1}+\tfrac{1}{3}k_1),\\
    k_3 & = h g_\phi(s_{l+1}+\tfrac{2}{3}h,v_{l+1}-\tfrac{1}{3}k_1+k_2),\\
    k_4 & = h g_\phi(s_l,v_{l+1}+k_1-k_2+k_3),\\
    v_l &= v_{l+1} + \tfrac{1}{8}(k_1+3k_2+3k_3+k_4).
    \end{split}
\label{eq:RK4}
\end{equation}



Generating matrix exponential coefficients $v_l,~l=1,2,..,L$ by the ODE solution (\ref{eq:euler}) or (\ref{eq:RK4}), the optimization for image registration (\ref{eqn:opt_me_mr}) using mutual information (\ref{eqn:MINE}) now becomes:
\begin{equation}
    \begin{split}
    \max_{\substack{u_1,u_2, \cdots \\ \phi, \theta}}
     \sum_{l=1}^L & \{MINE(T_l,Warp(M_l,Mexp(\sum_i v_{l,i} B_i)))+\\
     & MINE(M_l,Warp(T_l,Mexp(-\sum_i v_{l,i} B_i)))\}.
    \end{split}
\label{eqn:opt_me_mr2}
\end{equation}

\begin{figure}[h]
\centering
\includegraphics[scale=0.4]{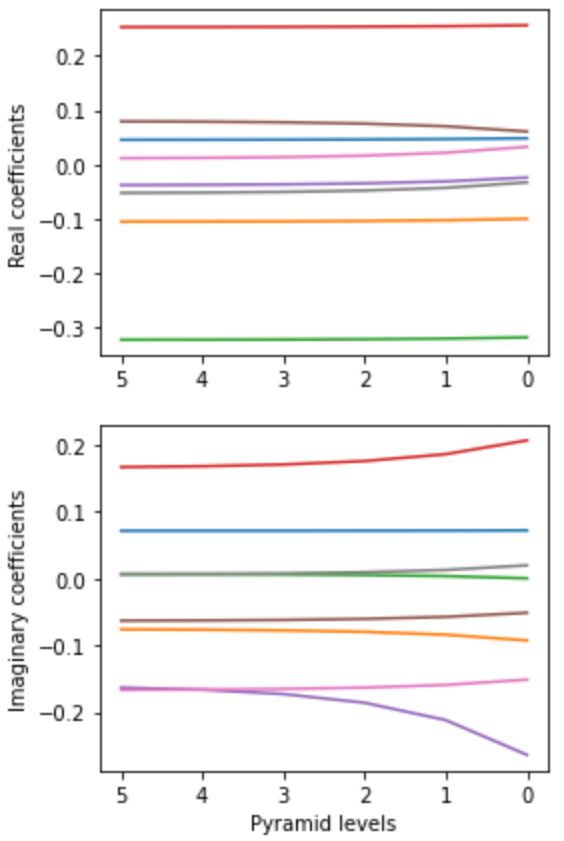}
\caption{Adaptation of matrix exponential coefficients over six levels for a registered image pair from ANHIR dataset. Level 5 is the coarsest resolution in the pyramid.}
\label{fig:imag_ode}
\end{figure}

The autograd feature of modern packages (e.g., PyTorch, Tensorflow) can easily work through the Euler or RK4 recursions for the optimization (\ref{eqn:opt_me_mr2}). Fig. \ref{fig:imag_ode} shows the adaptation of eight coefficients of complex matrix exponential over six resolution levels (0 being the original resolution). 

\subsection{Complex Matrix Exponential}

It is well known that exponential of real valued matrix is not globally surjective, i.e., not all transformation matrices (affine or homography) can be obtained by the exponential of real-valued matrices \cite{Gallier2020}. One way to overcome this issue is to compose matrix exponential a few times to compute the transformation matrix.

In this work, we propose to use complex matrix exponential an alternative to the scheme using composition, because complex matrix exponential is globally surjective \cite{Gallier2020}. Thus, a complex matrix, $B^r+\sqrt{-1}B^i = \sum_i v_i B_i$, produced by complex parameters, $v_i = v_i^r+\sqrt{-1}v_i^i$, can use matrix exponential series (\ref{eqn:mat_exp}) to create a complex transformation matrix,
\begin{equation}
    H^r + \sqrt{-1} H^i = Mexp(B^r + \sqrt{-1} B^i).
\label{eq:cme}
\end{equation}
Next, we choose to transform a point $(x,y)$ to another point $(x^\prime,y^\prime)$ using the following:
\begin{equation}
    \begin{split}
    [x^r,y^r,z^r]^T = H^r [x,y,1]^T,~ &  [x^i,y^i,z^i]^T = H^i [x,y,1]^T,\\
    x^\prime = \frac{x^r z^r + x^i z^i}{(z^r)^2+(z^i)^2},~ & 
    y^\prime = \frac{y^r z^r + y^i z^i}{(z^r)^2+(z^i)^2}.
    \end{split}
\label{eq:complex_hom}
\end{equation}

\begin{figure}[h]
\centering
\includegraphics[scale=0.2]{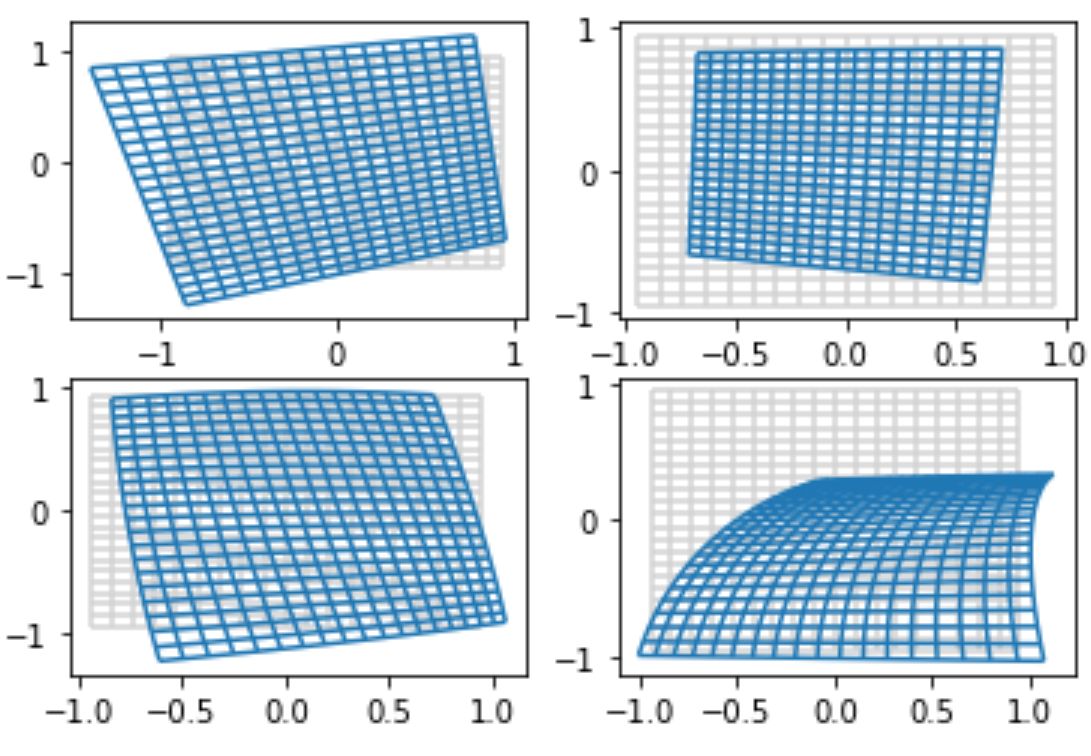}
\caption{Randomly generated grids by complex matrix exponential (\ref{eq:cme}) and complex transformation (\ref{eq:complex_hom}). Elements of $B^r$ were generated by a zero mean Gaussian with 0.1 standard deviation (SD) for all four panels. Elements of $B^i$ were generated by a zero mean Gaussian with SD as follows: 0 for top-left, 0.1 for top-right, 0.2 for bottom-left and 0.3 for bottom-right panel.}
\label{fig:grids}
\end{figure}

Note that under our chosen transformation (\ref{eq:complex_hom}) the straight lines are not guaranteed to remain straight. However, if $H^i=0$, transformation (\ref{eq:complex_hom}) degenerates to a linear transformation using homogeneous coordinates. Fig. \ref{fig:grids} shows four randomly generated grids using (\ref{eq:cme}) and (\ref{eq:complex_hom}). When imaginary coefficients are zeros (top-left panel, where $B^i=0$ and consequently $H^i=0$), the transformation acts as a homography, whereas the degree of the non-linearity in the transformation increases as the magnitude of $B^i$ increases. Unlike, a 2D Mobius transformation \cite{Kisil2012}, the proposed complex transformation does not guarantee self-intersection. However, note that 2D Mobius transformation is more restrictive, as for example, it cannot generate a perspective transformation.

\subsection{ODECME Algorithm}

Combining the aforementioned two elements, ordinary differential equation (ODE) and complex matrix exponential (CME), our proposed Algorithm \ref{alg:ODECME} (ODECME) first builds two image pyramids, one for the fixed and another for the moving image. It then computes Euler recursion for ODE-based computation of CME coefficients. Alternatively, we have also used RK4 recursion (\ref{eq:RK4}) in our experiments. Note also that ``Mexp'' may refer to real or complex matrix exponential, depending on whether $u$ and $v_l$ are complex or real. Also, ``MINE'' can be replaced by any differentiable loss for image registration.  $f_\theta$ (refer to (\ref{eqn:MINE})) is a fully connected neural network \cite{Nan2020}. $g_\phi$ is also a fully connected neural network  appearing in (\ref{eq:euler}) and (\ref{eq:RK4}). Taking advantage of matrix exponential, we use a symmetric loss, which uses both the forward and the inverse transformation matrices. For any gradient computation, such as $\nabla_{\theta}MI$ or $\nabla_{\phi}MI$, we use autograd (bult-in optimizers) of PyTorch \cite{NEURIPS2019_9015}. Algorithm \ref{alg:ODECME} finally outputs original resolution transformation matrix and its inverse.

\begin{algorithm}[h]
\SetAlgoLined
    Build multiresolution image pyramids $\{T_l,M_l\}_{l=1}^{L}$ \;
    Set learning rates $\alpha$, $\beta$ and $\gamma$\;
    Use random initialization for $\theta$ and $\phi$ \;
    Initialize $u$ to the 0 vector \;
    \For {each iteration}{
        $v_L = u$ \;
        $H_L = Mexp(\sum_i v_{L,i} B_i)$ \;
        $H_L^{-1} = Mexp(-\sum_i v_{L,i} B_i)$ \;
        \For {$l = [L-1,..,1]$}{
            $v_l = v_{l+1} + (s_l-s_{l+1})f_\phi(s_{l+1},v_{l+1})$ \;
            $H_l = Mexp(\sum_i v_{l,i} B_i)$ \;
            $H_l^{-1} = Mexp(-\sum_i v_{l,i} B_i)$ \;
        }
        $MI = 0$ \;
        \For {$l = [1,L]$}{
            $ MI \pluseq MINE(T_l, Warp(M_l, H_l))$ \;
            $ MI \pluseq MINE(M_l, Warp(T_l, H_l^{-1}))$ \;
        }
        Update parameter: $\theta \pluseq \alpha \nabla_{\theta} MI$ \;
        Update parameter: $\phi \pluseq \beta \nabla_{\phi} MI$ \;     Update parameter: $u \pluseq \gamma \nabla_u MI$ \;
    }
    Compute final transformation matrices:\\
    $v_L = u$ \;
    \For {$l = [L-1,..,1]$}{
        $v_l = v_{l+1} + (s_l-s_{l+1})f_\phi(s_{l+1},v_{l+1})$ \;
    }
    $H_1 = Mexp(\sum_i v_{1,i} B_i)$ \;
    $H_1^{-1} = Mexp(-\sum_i v_{1,i} B_i)$\;
\caption{ODECME}
\label{alg:ODECME}
\end{algorithm}

\section{Datasets for Experiments}
In order to evaluate our algorithm, we choose four datasets, two of them have 2D images: FIRE \cite{hernandez2017fire} and ANHIR \cite{ANHIR} and the other two are 3D volumes: IXI\cite{IXI} and ADNI \cite{wyman2013standardization}. FIRE and IXI are used to perform mono-modal registration, while ANHIR and ADNI are used for multi-modal registration.

\subsection{FIRE}
The FIRE dataset consists of 134 retinal fundus image pairs. These pairs are classified into three categories depending on what purpose they were collected for: S, P (Mosaicing) and A. Of these, for Category P pairs having $<75\%$ overlap, the registration optimization diverges in a lot of cases, and hence we leave out this subset of images in our experiments and use only Categories S and A. The FIRE dataset provides ground truth in the form of coordinates of 10 corresponding points between the fixed and the moving image. Also, while the images are square in shape, the retinal fundus is circular in shape and hence the gap between the edges of the fundus and the image border is quite large.So, we crop the central portion of the image to only include only the fundus ($1941 \times 1941$ pixels). For evaluating registration accuracy, we compute the Euclidean distance between these corresponding points after registration and average them. Also the image coordinates are scaled between 0 and 1 so that images of different sizes can be compared using the same benchmark. We call this measure the Normalized Average Euclidean Distance (NAED). Most competing methods do not support a homography based registration model, so an affine model was used to be consistent all across.

\subsection{ANHIR}
The ANHIR dataset provides high-resolution histopathological tissue images stained with different dyes. This provides a multi-modal challenge. The ground truth is provided in a similar format to the FIRE dataset. We use only the training set (230) provided in the database, since only these pairs have the ground truth available. The ANHIR dataset has some very large resolution images (upto $100k \times 200k$ pixels). Some of the competing registration frameworks were unable to process such large images and so, we downscaled every image by a factor of 5 to make them available to every framework. Furthermore, each staining can have a different resolution, so, to remedy this, we rescale the image with a smaller aspect ratio to match the width of the paired image and then it's height is padded to match the other image as well. This preprocessing is consistent across all algorithms and allows us to maintain the aspect ratios of the 
individual images and have both images in a pair at the same resolution. We use NAED as the evaluation metric here as well and use an affine model for transformation.

\subsection{IXI}
The IXI dataset has about 600 MR images from healthy patients. It includes T1, T2, PD-weighted images, MRA images, and Diffusion-weighted images. Unfortunately, the IXI dataset does not come with any form of ground truth, so we resort to standard measures \cite{ghosal2017deep} for registration accuracy such as SSIM and PSNR. We choose 51 T1 volumes (at random) which have the same size and designate one as the Atlas (reference volume) and register the other 50 volumes against it. The SSIM and PSNR scores are computed after every registration with the Atlas and averaged and then reported for each algorithm. The transformation model used has 7 degrees of freedom: isotropic scaling with three axes of rigid transformation and three axes for rotation.

\subsection{ADNI}
The ADNI dataset provides 1.5T and 3T MRI scans of patients scanned over different periods of time. We chose one volume as the atlas (template volume) from the ADNI1:Screening 1.5T collection and another 50 volumes from the ADNI1:Baseline 3T collection. All volumes were normalized, and resized to match the reference volume ($160 \times 192 \times 192$) before feeding into any of the algorithms. Similar to the IXI dataset, no ground truth is available here for registration, so we report the averaged SSIM and PSNR metrics for the dataset after registration. The transformation model is the same as the one used for IXI.

\begin{figure}[h!]
    \centering
    \includegraphics[scale=0.3]{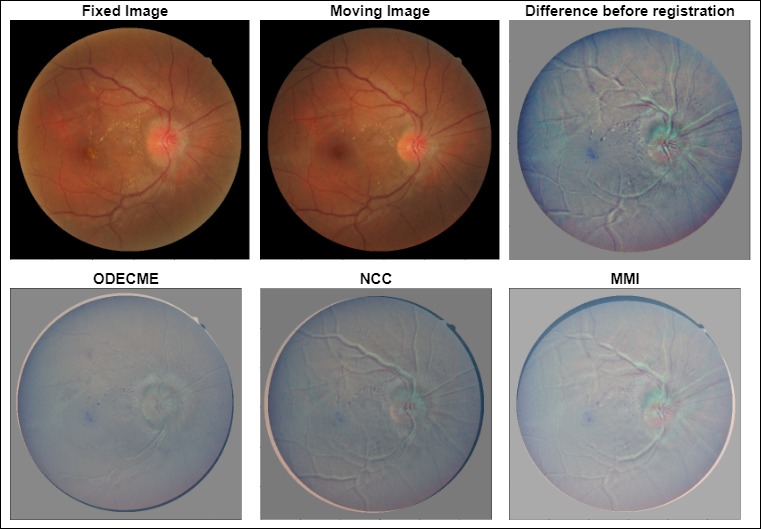}
    \caption{Visual results for a pair of registered images from the FIRE dataset. Bottom row shows difference images after registration with the best three algorithms. Here ODECME refers to the RK4-Complex version.}
    \label{fig:fire_samples}
\end{figure}

\section{Experiments}
Algorithm \ref{alg:ODECME} with real matrix exponential and without multi-resolution adaptation by ODE has been published as DRMIME \cite{Nan2020}. We compare our proposed enhanced version ODECME with DRMIME and other competing methods for all four datasets. For all algorithms and datasets, we use $L=6$ for the maximum level of Gaussian image pyramid. We use $\alpha=0.1, \beta=\gamma=0.01$ for the 2D datasets and $\alpha=0.01, \beta=\gamma=0.001$ for the 3D datasets in Algorithm \ref{alg:ODECME}. 

We use MINE as the objective function for all four datasets. Our implementation of the network $f_\theta$ for MINE uses a fully connected network with twice the number of input channels as the input layer, e.g., for a color image it is $3 \times 2 = 6$. There are two hidden layers with 100 neurons in each and the output layer has a scalar output. Apart from the output layer which has no activation, ReLU activation is used. 

For ODE, the input layer for $g_\phi$ consists of $7$ and $8$ neurons in case of FIRE/ANHIR and IXI/ADNI, respectively. The reasoning being, that one neuron accounts for the scale of the level in the Gaussian pyramid and remaining neurons are for the number of matrix exponential coefficients ($6$ for 2D and $7$ for 3D datasets in our experiments). For complex coefficients, these input dimensions become $13$ and $17$, respectively. $g_\phi$ has a single hidden layer with ReLU activation and $100$ neurons and the final output layer consists of neurons equal to the number of matrix exponential coefficients.

For all evaluations, we also conduct a paired t-test with DRMIME to investigate if the results are statistically significant (p-value $<$ 0.05).

\subsection{Competing Methods}
We evaluate our method against the following off-the-shelf registration algorithms from popular registration frameworks. For a fair comparison, we perform random grid search to set various hyperparameters of these toolboxes. A detained description of these hyper parameters can be found in DRMIME \cite{Nan2020}. We compare ODECME with the following methods:
\begin{enumerate}
    \item Mattes Mutual Information (MMI) \cite{mattes2001nonrigid, mattes2003pet, MMI}
    
    \item Joint Histogram Mutual Information (JHMI) \cite{thevenaz2000optimization, JHMI}
    \item Normalized Cross Correlation (NCC)\cite{NCC}
    \item Mean Square Error (MSE)\cite{MSE}
    
    \item AirLab Mutual Information (AMI)\cite{DBLP:journals/corr/abs-1806-09907}
    \item Normalized Mutual Information (NMI)\cite{studholme1999overlap, NMI}, and
    \item DRMIME \cite{Nan2020}
\end{enumerate}

The implementations of the above algorithms were used from these packages:
\begin{itemize}
    \item SITK: MMI, JHMI, NCC, MSE
    \item AirLab: AMI
    \item SimpleElastix: NMI
\end{itemize}

\subsection{Accuracy Comparisons}
\begin{figure}[h]
\centering
\includegraphics[scale=0.4]{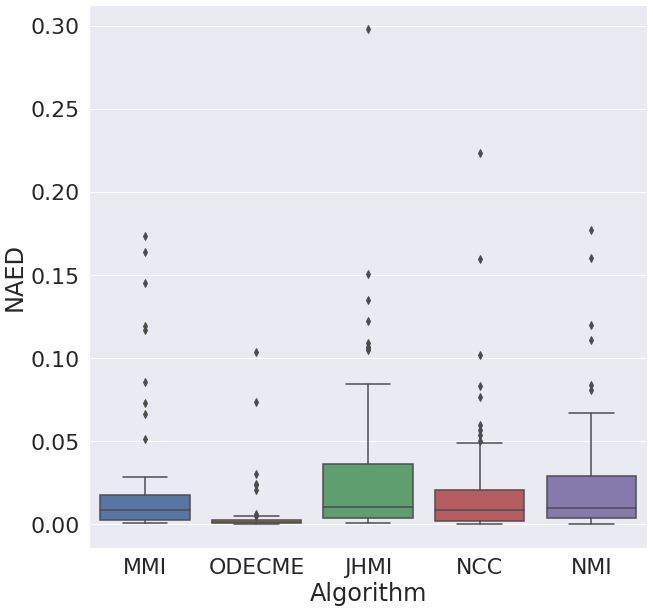}
\caption{Box plot for NAED of the best 5 performing algorithms on FIRE}
\label{fig:fire_res}
\end{figure}

Fig. \ref{fig:fire_samples} shows registration results for a randomly chosen image pair from FIRE dataset. Table \ref{tab:fire_res} shows the NAED for all algorithms on the FIRE dataset. We observe that Runge-Kutta ODE recursion with complex matrix exponential, ODE (RK4-Complex), performs significant better than the competitors including DRMIME. We also note that Runge-Kutta version is more effective than the Euler version. Complex version did not have any significant advantage over the real version for accuracy. Fig. \ref{fig:fire_res} presents box plots for ODE (RK4-Complex) and results from four other toolboxes. We notice that number of outliers is the lowest in ODECME illustrating its robustness.

    \begin{table}[t]
    \caption{NAED for FIRE dataset along with paired t-test significance values}
    \centering
    \begin{tabular}{||c|c|c||}
        \hline
        Algorithm & NAED (Mean $\pm$ STD) & p-value \\ 
         \hline\hline
         ODE (RK4-Complex) & \textbf{0.00380} $\pm$ 0.012 & 0.0032 \\
         \hline
         ODE (RK4-Real) & 0.00385 $\pm$ 0.014 & 0.0032 \\
         \hline
         ODE (Euler-Complex) & 0.0047 $\pm$ 0.019 & 0.0822 \\ 
         \hline
         ODE (Euler-Real) & 0.0049 $\pm$ 0.016 & 0.1053 \\ 
         \hline
         DRMIME (Complex) & 0.00482 $\pm$ 0.031 & 0.1021 \\ 
         \hline
         DRMIME (Real) & 0.00482 $\pm$ 0.026 & - \\ 
         \hline
         NCC & 0.0194 $\pm$ 0.033 & 1.3e-04 \\
         \hline
         MMI & 0.0198 $\pm$ 0.034 & 5.4e-05 \\
         \hline
         NMI & 0.0228 $\pm$ 0.032 & 1.7e-08 \\
         \hline
         JHMI & 0.0311 $\pm$ 0.046 & 4.5e-07 \\
         \hline
         AMI & 0.0441 $\pm$ 0.028 & 1.4e-27 \\
         \hline
         MSE & 0.0641 $\pm$ 0.094 & 3.5e-03 \\ [1ex] 
        \hline
    \end{tabular}
    \label{tab:fire_res}
    \end{table}

\begin{figure*}[h]
    \centering
    \includegraphics[scale=0.4]{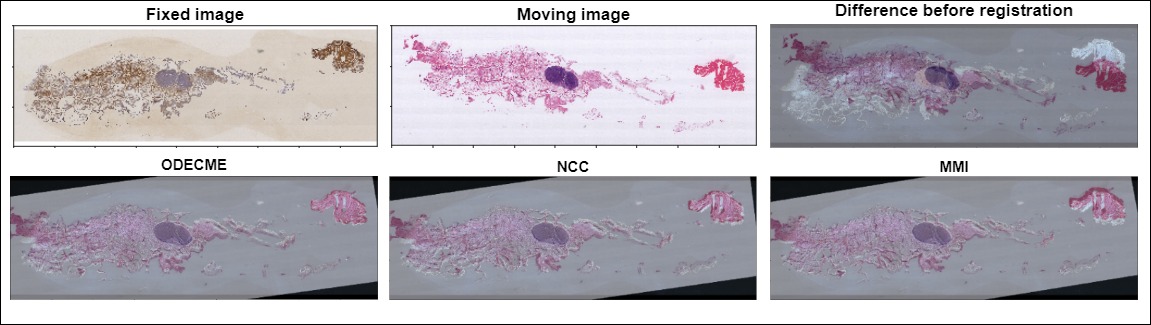}
    \caption{Visual results for a pair of registered images from the ANHIR dataset. Bottom row shows difference images after registration with the best three algorithms. Here ODECME refers to the RK4-Complex version.}
    \label{fig:anhir_samples}
\end{figure*}

\begin{figure}[h]
\centering
\includegraphics[scale=0.4]{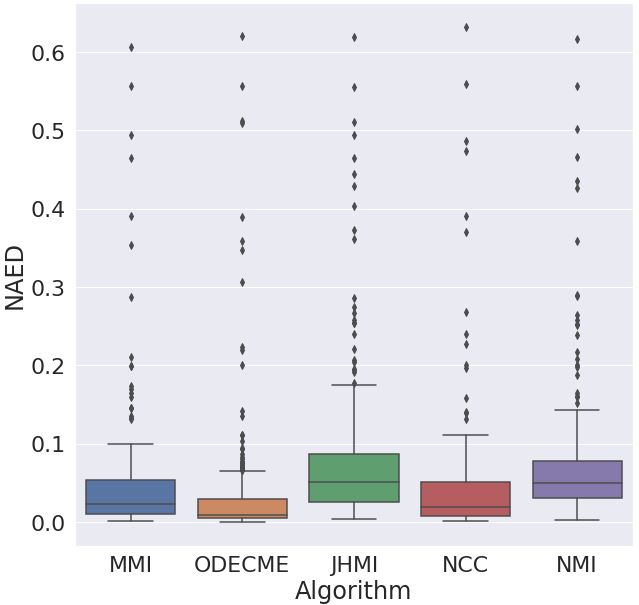}
\caption{Box plot for top 5 performing algorithms on ANHIR}
\label{fig:anhir_res}
\end{figure}

Fig. \ref{fig:anhir_samples} shows registration results for a sample image pair from ANHIR dataset. Table \ref{tab:anhir_res} presents the NAED metrics, where once again we notice that ODE (RK4-Complex) produced the best accuracy. Similar to the FIRE dataset, Runge-Kutta method produced better results than the Euler recursion. For accuracy, as before, we have not spotted any obvious advantage of complex coefficients over the real ones. The box-plots in Fig. \ref{fig:anhir_res} also emphasise the same conclusion as we saw before, i.e. ODECME outperforms other competing algorithms.

\begin{table}[h]
    \caption{NAED for ANHIR dataset along with paired t-test significance values}
    \centering
    \begin{tabular}{||c|c|c||}
        \hline
        Algorithm & NAED (Mean $\pm$ STD) & p-value \\ 
         \hline\hline
         ODE (RK4-Complex) & \textbf{0.0344} $\pm$ 0.045 & 0.0441 \\
         \hline
         ODE (RK4-Real) & 0.0348 $\pm$ 0.044 & 0.0642 \\
         \hline
         ODE (Euler-Complex) & 0.0358 $\pm$ 0.075 & 1.0e-03 \\ 
         \hline
         ODE (Euler-Real) & 0.0391 $\pm$ 0.035 & 1.5e-03 \\ 
         \hline
         DRMIME (Complex) & 0.0373 $\pm$ 0.021 & 0.0619 \\ 
         \hline
         DRMIME (Real) & 0.0373 $\pm$ 0.015 & - \\ 
         \hline
         NCC & 0.0461 $\pm$ 0.084 & 7.0e-04 \\
         \hline
         MMI & 0.0490 $\pm$ 0.082 & 6.2e-05 \\
         \hline
         MSE & 0.0641 $\pm$ 0.094 & 5.5e-14 \\
         \hline
         NMI & 0.0765 $\pm$ 0.090 & 3.0e-31 \\
        \hline
         AMI & 0.0769 $\pm$ 0.090 & 3.7e-30 \\
         \hline
        JHMI & 0.0827 $\pm$ 0.100 & 8.3e-21 \\  [1ex] 
         \hline
    \end{tabular}
    \label{tab:anhir_res}
\end{table}

\begin{figure}[h]
    \centering
    \includegraphics[scale=0.4]{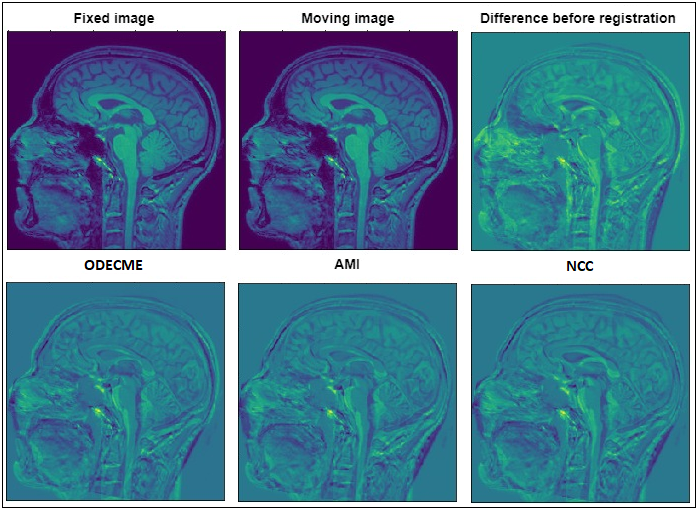}
    \caption{Difference images from the middle slice from a pair of volumes from the IXI dataset shown before and after registration with three top performing algorithms. ODECME refers to the RK4-Complex version.}
    \label{fig:ixi_samples}
\end{figure}

\begin{figure}[h]
\centering
\includegraphics[scale=0.4]{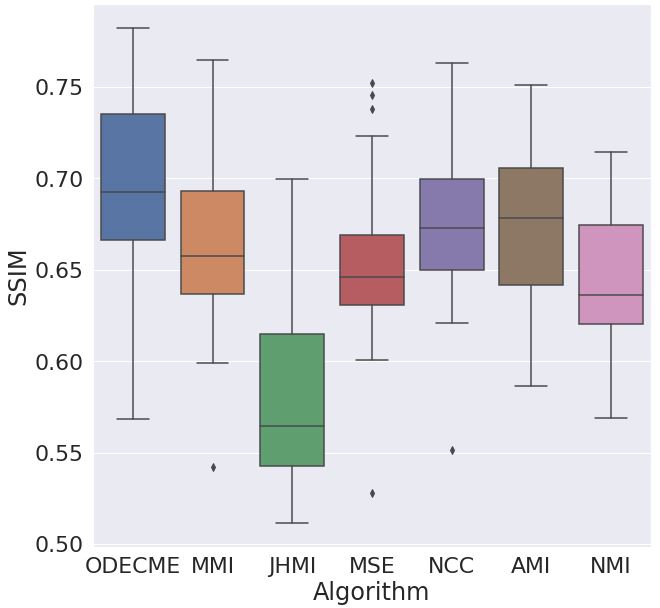}
\caption{Box plot for SSIM values for each algorithm on the IXI datset after registration. ODECME refers to the RK4-Complex version.}
\label{fig:ixi_ssim}
\end{figure}

\begin{figure}[h]
\centering
\includegraphics[scale=0.4]{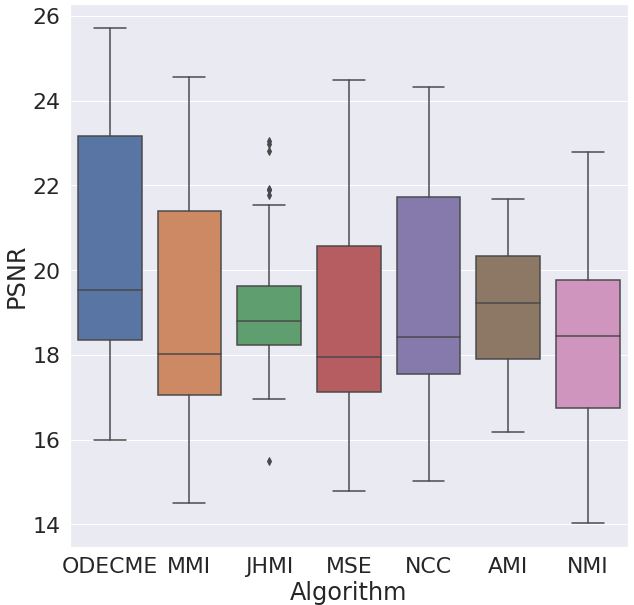}
\caption{Box plot for PSNR values for each algorithm on the IXI datset after registration. ODECME refers to the RK4-Complex version.}
\label{fig:ixi_psnr}
\end{figure}

Fig. \ref{fig:ixi_samples} shows sample results for IXI dataset with three top performing algorithms. We compute the SSIM and PSNR scores after registration and present them in Fig. \ref{fig:ixi_ssim} and \ref{fig:ixi_psnr}, respectively. Both these figures show that ODE (RK4-Complex) is significantly better.

\begin{figure}[h]
\centering
\includegraphics[scale=0.4]{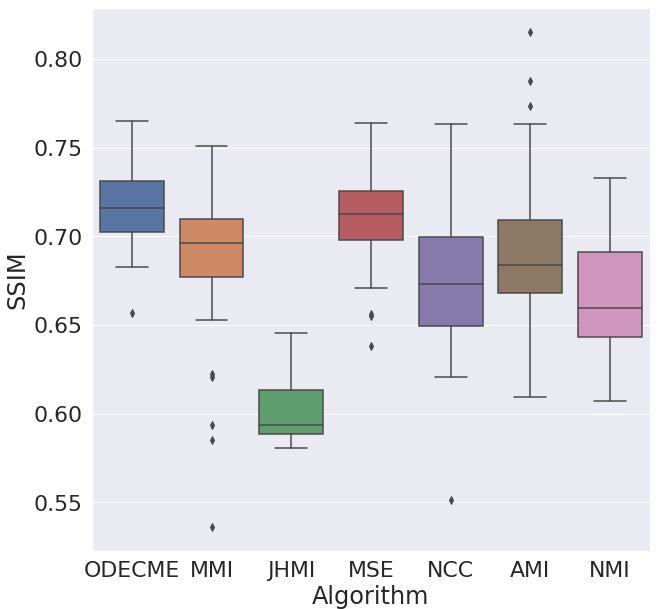}
\caption{Box plot for SSIM values for each algorithm on the ADNI dataset after registration. ODECME refers to the RK4-Complex version.}
\label{fig:adni_ssim}
\end{figure}

\begin{figure}[h]
\centering
\includegraphics[scale=0.4]{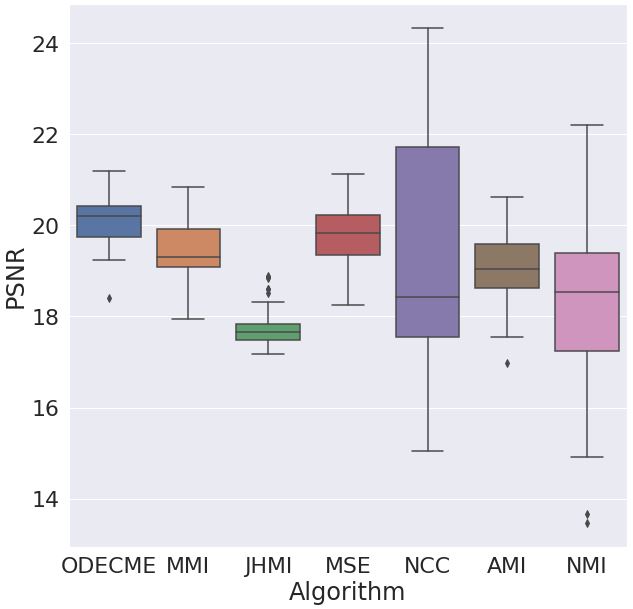}
\caption{Box plot for PSNR values for each algorithm on the ADNI dataset after registration. ODECME refers to the RK4-Complex version.}
\label{fig:adni_psnr}
\end{figure}

\begin{figure*}[h]
    \centering
    \includegraphics[scale=0.40]{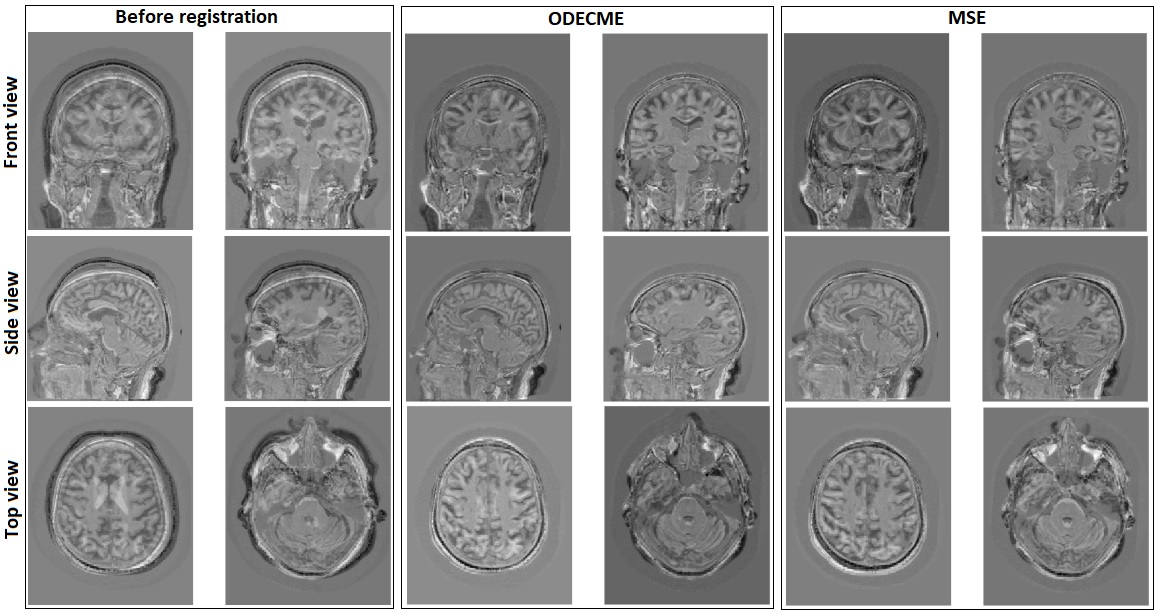}
    \caption{Same slices from the difference volume after before and after registration with the two top performing algorithm.}
    \label{fig:ADNI_samples}
\end{figure*}

We plot SSIM and PSNR scores for ADNI dataset in Figures \ref{fig:adni_ssim} and \ref{fig:adni_psnr}, respectively. For ODECME the median scores not only are higher, but also produced the lowest spread/range. Fig. \ref{fig:ADNI_samples} shows Two different slices with three views before and after registrations with ODECME and MSE algorithms, which according to our experiments, is the next best algorithm for this dataset.

\subsection{Effect of CME}
\begin{figure}[h]
\centering
\includegraphics[scale=0.3]{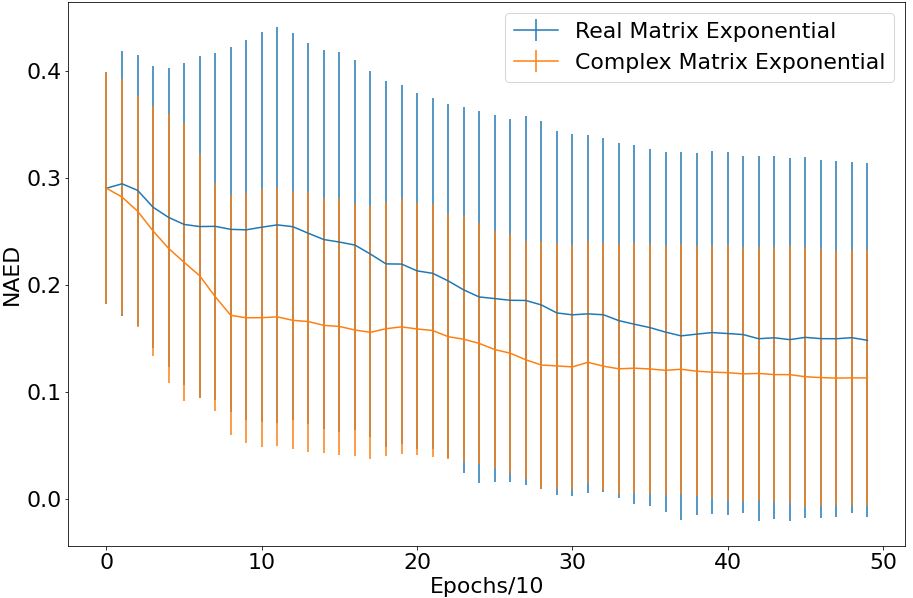}
\caption{NAED averaged for 10 randomly selected pairs from FIRE, plotted over 500 epochs. Error bars represent the range of NAED values over 10 registrations.}
\label{fig:compvsreal}
\end{figure}

While the NAED performance results are not statistically significant to be able to conclude better accuracy for CME, it speeds up convergence. For instance, with real matrix exponential, for the FIRE dataset, it takes about 500 epochs to converge, while with ANHIR it takes about 1500 epochs. In case of complex matrix exponential, it only takes about 300 epochs in case of FIRE, and about 1300 epochs for ANHIR. Fig. \ref{fig:compvsreal} shows the NAED convergence graphs (error bar plots) for 10 randomly selected pairs from FIRE for Algorithm \ref{alg:ODECME} using both real and complex matrix exponential coefficients. The plots show that convergence using CME is much better.

\subsection{Effect of ODE}
To demonstrate that Algorithm \ref{alg:ODECME} can fine-tune transformation matrices over the resolution levels, we compute the range of each complex coefficient after registering FIRE dataset over six resolution levels. Table \ref{tab:ode_avg_std} shows the average and standard deviations of these range values over the entire FIRE dataset. Note that the transformation matrices act over the canonical pixel value range of $[-1,1]\times[-1,1].$ Thus, the  small variations of the matrix exponential coefficients are still significant changes for the transformation matrices. Also note the contribution from imaginary coefficients are significant in designing the transformation matrices. 

\begin{center}
    \begin{table}[h]
    \caption{The average range and standard deviation of real and imaginary coefficients after registering the FIRE dataset}
    \centering
    \begin{tabular}{||c|c|c||}
        \hline
        Coefficient & (Real) Mean Range $\pm$ SD & (Imag.) Mean Range $\pm$ SD \\ 
         \hline\hline
         $v_{0}$ & 0.0107 $\pm$ 0.0100 & 0.0665 $\pm$ 0.0605\\
         \hline
         $v_{1}$ & 0.0119 $\pm$ 0.0074 & 0.0188 $\pm$ 0.0116\\ 
         \hline
         $v_{2}$ & 0.0400 $\pm$ 0.0122 & 0.0637 $\pm$ 0.0371\\ 
         \hline
         $v_{3}$ & 0.0233 $\pm$ 0.0177 & 0.0299 $\pm$ 0.0222\\
         \hline
         $v_{4}$ & 0.0212 $\pm$ 0.0107 & 0.0654 $\pm$ 0.0407\\
         \hline
         $v_{5}$ & 0.0135 $\pm$ 0.0052 & 0.0892 $\pm$ 0.0624\\
         \hline
    \end{tabular}
    \label{tab:ode_avg_std}
\end{table}
\end{center}

\subsection{Running Time}
To provide the running times of all the algorithms, we choose FIRE dataset and run all algorithms for $1000$ iterations on a desktop computer with a single NVIDIA GeForce GTX 1080 Ti, Intel(R) Xeon(R) CPU E5-2620 v4 @ 2.10GHz, 32GB RAM. Table \ref{tab:perf_table1} shows running time. We note that except ODECME, DRMIME, and AMI, all other algorithms did not use GPU accelerations. Additionally, we ran ODECME (RK4) for 50 epochs that resulted in an average NAED, which is better than the most competitors for a similar running time. Note that our software is not optimized, unlike SimpleElastix (NMI) for example. 

\begin{center}
    \begin{table}[h]
    \caption{Time taken for 1000 epochs and resultant NAED (lower is better)}
    \centering
    \begin{tabular}{||c|c c||}
        \hline
        Algorithm & Time (seconds) & NAED \\ 
         \hline\hline
         ODECME (RK4) (50 epochs) & 108 & 0.01921\\
         \hline
         NMI & 60 & 0.02503\\
         \hline
         AMI & 620 & 0.02942\\ 
         \hline
         DRMIME & 1425 & 0.00368 \\ 
         \hline
         ODECME (RK4) & 1601 & \textbf{0.00360}\\
         \hline
         ODECME (Euler) & 1469 & 0.00452\\
         \hline
         MMI & 2904 & 0.00598\\
         \hline
         JHMI & 1859 & 0.00605\\
         \hline
         NCC & 3804 & 0.00697\\
          \hline
         MSE & 2847 & 0.02918\\ [1ex] 
        \hline
    \end{tabular}
    \label{tab:perf_table1}
\end{table}
\end{center}

\section{Conclusion and Future Work}
Optimization-based image registration using homography as a transformation is classical. However, recent toolboxes with autograd capability and strong GPU acceleration has created opportunity to improve these classical registration algorithms. In this work, we show that using complex matrix exponential convergence can be accelerated for such algorithms. Also using ordinary differential equation, we can further refine the accuracy of such algorithms for multi-resolution image registration problems that is able to employ any differentiable objective function. Our algorithm yields state-of-the-art accuracy for benchmark 2D and 3D datasets. We plan to employ the ODE framework for deformable registration in our future endeavor.

\noindent \textbf{Author Roles:} The first author wrote the software for this work, conducted experiments, and produced results and contributed in writing. Other authors supervised the first author. The last author contributed in conceptualizing and writing.

\bibliography{bibfile}
\bibliographystyle{IEEEtran}

\end{document}